\documentclass[11pt]{article}

\usepackage[margin=1in]{geometry}
\usepackage[utf8]{inputenc}
\usepackage[T1]{fontenc}
\usepackage{newtxtext,newtxmath}
\usepackage{amsmath,amstext}
\usepackage{graphicx}
\usepackage{booktabs}
\usepackage{longtable}
\usepackage{array}
\usepackage{calc}
\usepackage{etoolbox}
\usepackage{caption}
\usepackage{float}
\usepackage{natbib}
\usepackage{hyperref}
\usepackage{bookmark}
\usepackage[capitalize,noabbrev]{cleveref}
\usepackage{parskip}

\hypersetup{
    colorlinks=true,
    linkcolor=blue,
    citecolor=blue,
    urlcolor=blue
}

\title{Marginal Advantage Accumulation for Memory-Driven Agent Self-Evolution}
\author{
    Mingyu Yang, Keye Zheng, Congchao Cheng$^\dagger$, Yujie Liu, Xingkang Lu \\
    Fan Jiang$^\dagger$, Yefei Zheng$^\dagger$ \\[6pt]
    Alibaba International Digital Commerce Group \\
    \texttt{\{mingyu.yang, zhengkeye.zky, congchao.chengcong\}@alibaba-inc.com} \\
    \texttt{\{lyj315363, luxingkang.lxk, zhanyan, yuanxin\}@taobao.com} \\[3pt]
    $^\dagger$Corresponding authors.
}
\date{}

\begin{document}
\maketitle

\begin{abstract}
In batch-style trace distillation, the same memory operation may receive contradictory feedback across different batches. Existing methods lack a cross-batch, operation-level evidence accumulation mechanism, making it impossible to distinguish stably effective operations from accidental hits. This paper formalizes the requirement as two structural conditions, alignability and comparability, and proposes Marginal Advantage Accumulation (MAA). MAA constructs differential signals to make them comparable across batches, accumulates signed evidence per operation via EMA, and ensures cross-batch traceability through semantic identity merging. As a post-processing architecture, MAA achieves the best results in 14 out of 16 settings across 4 benchmarks and 4 target models, consistently outperforming existing batch-level distillation baselines and matching or surpassing online alternatives in most settings, while reducing optimization-phase token consumption by approximately 75\%.

\end{abstract}

\section{Introduction}

\subsection{Background: Agent Self-Evolution and Trace Distillation}

As LLM-driven autonomous agents take on increasingly complex responsibilities in scientific discovery \cite{chen2025scienceagentbench}, embodied interaction \cite{shridhar2021alfworld}, engineering, and everyday tasks, a central question comes to the fore: how can agents accumulate reusable capability improvements during continuous use? This problem is commonly referred to as agent self-evolution.

Within the non-parametric self-evolution paradigm, trace distillation occupies a central position. Regardless of whether the final memory takes the form of skills \cite{wang2024voyager}, experiential insights \cite{zhao2024expel}, workflow templates \cite{wang2025awm}, retrieval cases, or few-shot examples, all methods face the same subtask: compressing a set of execution traces $\{\tau_1, \dots, \tau_N\}$ into a reusable memory bank $M = \{m_1, \dots, m_K\}$. The distilled $M$ is injected into the LLM context at inference time, directly affecting subsequent decision quality. Distillation quality therefore sets the performance ceiling of the non-parametric self-evolution paradigm.

\subsection{Design Gaps in Batch-Style Trace Distillation}

The typical scenario in industrial deployment is as follows: an agent first executes on a batch of tasks and accumulates traces, after which the system optimizes memory from these traces without re-rollout. This setting imposes two hard constraints. First, the ``per-step re-evaluation'' required by online distillation is impractical---a full agent execution involves multiple rounds of LLM calls and tool interactions, making re-rollout costs typically an order of magnitude higher than distillation itself. Assessing suggestion quality can only rely on cheap proxy signals (e.g., an LLM judge) rather than true forward verification. Second, trace collections far exceed a single LLM context window, forcing distillation to proceed in units of batch $B_t$---a choice shared by existing methods such as SkillOpt \cite{yang2026skillopt} and Trace2Skill \cite{ni2026trace2skill}.

The combination of these two constraints restricts optimization decisions to the local perspective of a single batch. Consider scientific data analysis as an example: an agent exposes ``plotting code missing axis labels causing review rejection'' in failed traces of one batch, and the system accordingly proposes ``automatically append \texttt{plt.xlabel} / \texttt{plt.ylabel} at the end of all plotting code.'' This suggestion is beneficial in the current batch (all tasks involve standard 2D charts), but in another batch targeting heatmaps or 3D scatter plots, forcibly adding 2D axis labels leads to rendering errors instead. Such ``locally effective, globally unstable'' operations are pervasive in batch-style distillation: proxy signals are inherently noisy, the same suggestion may receive contradictory feedback across different batches, and making accept-or-reject decisions based solely on single-batch local perspectives prevents the system from leveraging subsequent reverse signals to correct earlier judgments---thus failing to distinguish ``stably effective'' from ``accidentally hit.''

The essence of the problem is that existing methods lack a cross-batch evidence accumulation mechanism. To fill this design gap, an accumulation mechanism must satisfy two structural requirements:

\begin{enumerate}
\item \textbf{Cross-batch identity alignment (Alignability)}: The system must be able to identify semantically equivalent operations across different batches and merge them into the same accumulation unit. Without stable tracking of the same operation, cross-batch evidence accumulation is impossible.
\item \textbf{Cross-batch comparability (Comparability)}: Different batches have varying task distributions; the signals produced by the same operation across batches must share a consistent scale to enable meaningful additive accumulation. If signals are dominated by batch-specific characteristics rather than the true effect of the operation, cross-batch aggregation becomes meaningless.
\end{enumerate}

These two requirements reveal the fundamental distinction between step-local edit selection (ordinal selection on the current batch) and cross-batch edit accumulation (stateful evidence aggregation across batches): ranking or preference signals suffice for the former but cannot support the latter due to the absence of the above two properties. No existing method provides a design layer satisfying both requirements, and this deficiency worsens with increasing task complexity---when task distribution differences are large and trajectories are long, the value of cross-batch evidence accumulation becomes particularly significant.

\begin{figure}[t]
\centering
\includegraphics[width=\linewidth]{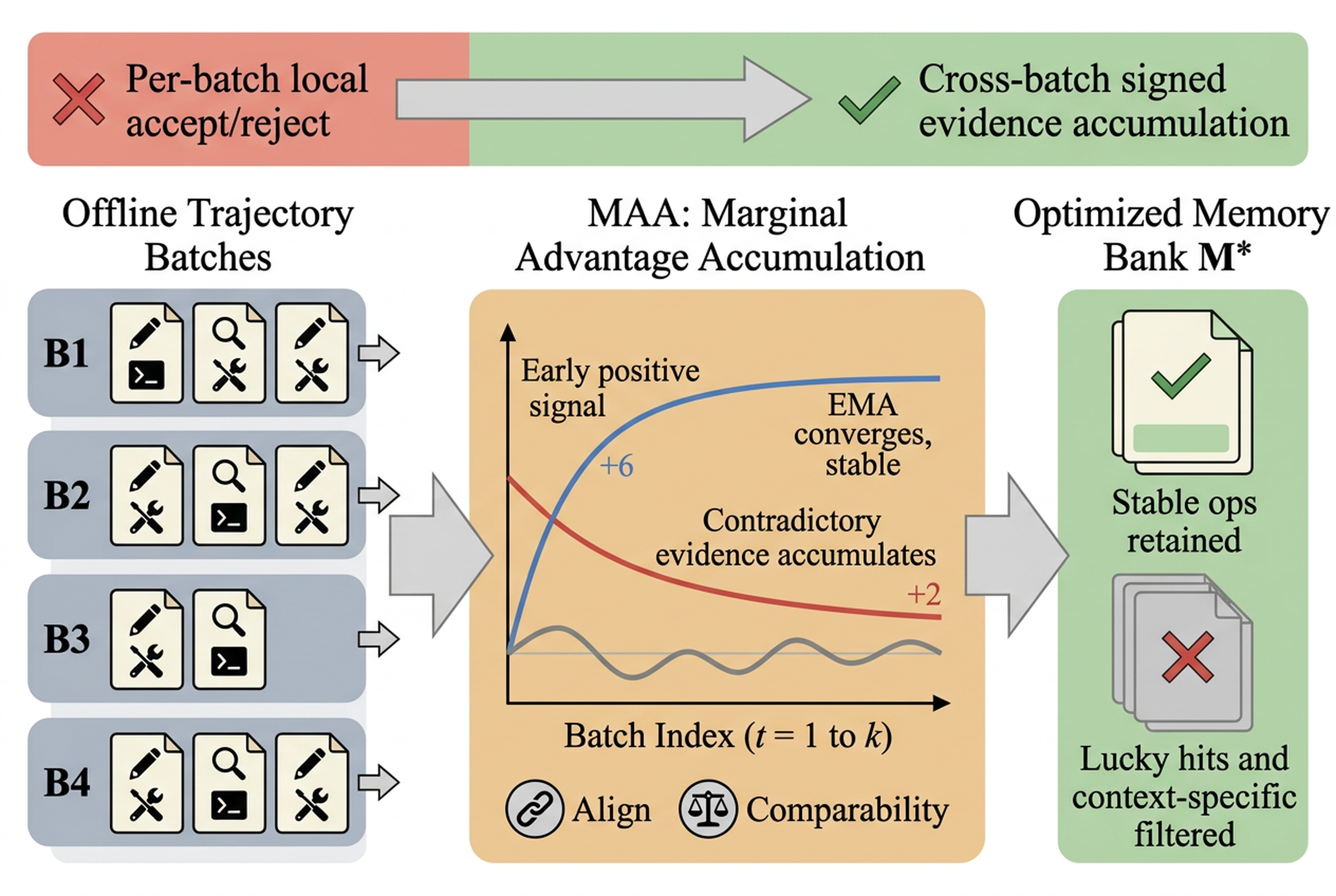}
\caption{Batch-style trace distillation problem and MAA solution. Single-batch methods cannot distinguish stable effects from accidental hits; MAA aggregates cross-batch evidence via identity merging, differencing, and EMA.}
\label{fig:problem_solution}
\end{figure}

\subsection{Method Overview and Contributions}

To address the above gap, we propose Marginal Advantage Accumulation (MAA) for Memory-Driven Agent Self-Evolution. As illustrated in Figure~\ref{fig:problem_solution}, the core idea of MAA is to ground user-perceived ``optimization suggestions'' into addressable atomic operations (ops) with invariant identity across batches, and on this basis construct scale-consistent accumulation signals so that evidence of the same op across different batches can be additively aggregated.

MAA uses the LLM to estimate the expected utility $u(M, B) \in [0, 100]$ (integer percentage) of the agent on the current batch $B$ given memory $M$, and performs outer-layer differencing for each candidate op $i$: $\delta_i = u(M_i, B) - u(M, B) \in [-100, 100]$. Differencing mitigates distribution differences between batches, making signals satisfy comparability; stable ids in the memory bank enable tracking the same op across batches, satisfying alignability. What is accumulated is not batch scores or one-time accept/reject judgments, but cumulative evidence of the same op across multiple batches. $u$ itself need not be an unbiased estimate of true task gain; MAA relies only on the weaker directional alignment, i.e., the direction of $\delta$ points toward the true gain direction with probability statistically better than random.

The method consists of an infrastructure layer and a core design layer:

\begin{itemize}
\item \textbf{Addressable memory bank and semantic identity merging} (\S3.1): The memory bank is abstracted as a set with stable item ids, where each op is identified by (type, anchor id, content embedding). When a new candidate's semantic similarity to an existing op in the pool exceeds a threshold, it is merged into the same accumulation unit, avoiding identity fragmentation caused by wording differences. This layer satisfies the alignability requirement.
\item \textbf{Marginal advantage and per-op cross-batch accumulation} (\S3.2--3.3): Baseline and candidate states are evaluated side-by-side, followed by outer-layer differencing, giving each op an incremental signal $\delta_i$ relative to the current baseline (satisfying comparability). Multi-step $\delta_{k,t}$ on each op unit $k$ is then aggregated via EMA, enabling three types of operations to be distinguished under multi-batch sampling: stably effective (EMA at high positive values), spurious correlation (EMA initially positive then trending to zero), and scene-specific (EMA oscillating around zero). Ablation experiments confirm that both differencing and accumulation layers jointly account for the performance gains.
\end{itemize}

Engineering governance (top-$k$ sparse truncation, best-checkpoint rollback, candidate pool management) serves as a supporting layer that, together with the method itself, forms the complete system.

Core contributions:

\begin{enumerate}
\item \textbf{Identification and formalization of design gaps}: We identify the missing cross-batch operation-level evidence accumulation design layer in batch-style trace distillation and formalize it as two structural requirements: alignability and comparability. This framework also applies to analyzing and improving future non-parametric self-evolution methods.
\item \textbf{MAA method}: We propose marginal advantage accumulation as a natural instantiation of the above design layer, satisfying the two requirements through semantic identity merging, differencing, and per-op EMA respectively. Each component is standard technology; their combination follows directly from the problem structure.
\item \textbf{Performance and efficiency}: MAA achieves the best results in 14 out of 16 settings across four benchmarks $\times$ four target models, reducing optimization-phase token consumption by approximately 75\%.
\end{enumerate}

\section{Related Work}

Agent self-evolution methods can be divided into online and offline categories based on whether online environment interaction is required. MAA belongs to the offline architecture; below we review related technical routes and their limitations within both frameworks.

\subsection{Offline Trace Distillation}

\textbf{Single-shot distillation.} Reflexion \cite{shinn2023reflexion}, Self-Refine \cite{madaan2023selfrefine}, ExpeL \cite{zhao2024expel}, Voyager \cite{wang2024voyager}, and AWM \cite{wang2025awm} extract reusable memory in a one-shot manner from the perspectives of verbal feedback, experiential insights, skill libraries, and workflows respectively. These methods perform distillation only once; once memory is written, it cannot be undone. They naturally sidestep alignability and comparability problems, but consequently cannot leverage multi-batch evidence cancellation to filter spurious operations---once an erroneous suggestion is written, subsequent batches cannot correct it even if they produce reverse signals.

\textbf{Offline hierarchical distillation.} Trace2Skill \cite{ni2026trace2skill} introduces hierarchical induction on top of single-shot distillation, progressively merging trajectory-local patches into global skills. Compared to single-shot distillation, it better abstracts cross-task common patterns, but still performs one-shot induction in batch units without retaining per-op decision evidence for subsequent batch correction. This means when an induced skill performs poorly on a new batch, the system lacks mechanisms to trace back and adjust its source evidence.

\textbf{Reactive evolvable memory.} A-MEM \cite{xu2025amem}, Evo-Memory \cite{wei2025evomemory}, MEMO \cite{xie2026memo}, MemGPT \cite{packer2024memgpt}, HyMEM \cite{zhu2026hymem}, and MemSkill \cite{zhang2026memskill} equip memory systems with update primitives such as add/edit/remove, supporting iterative evolution. Live-Evo \cite{zhang2026liveevo} further extends to online streaming scenarios, inducing rule-level memory from continuous feedback. Although these methods support memory updates, each decision is based only on the local perspective of the current batch and does not maintain op-level cross-batch identity. The consequence is that the same operation may be treated as different entities across batches, preventing aligned accumulation of positive and negative evidence. Meanwhile, batch difficulty effects entangled in score fluctuations cannot be disentangled, causing ``operation quality'' and ``batch difficulty'' to be confounded in decision-making.

\subsection{Online Optimization Methods}

SkillOpt \cite{yang2026skillopt} represents the online architecture: each optimization step deploys the agent to execute complete trajectories in the environment (online rollout), iteratively rewriting skill documents based on real feedback; SkillGrad \cite{wang2026skillgrad} follows a similar approach. Online methods obtain signals from real environments with high directional reliability, but each rollout step costs an order of magnitude more than distillation itself, making scaling difficult. MAA diverges from online methods along the computational paradigm axis: it operates only on existing traces, requiring no environment interaction during the optimization phase.

\subsection{Iterative Optimization Paradigms}

The online and offline methods above each adopt different optimization paradigms: single-shot distillation and hierarchical induction are one-shot extraction, while reactive memory follows stepwise responsive updating. Iterative optimization paradigms borrow the formal discipline of parameter optimization to organize multi-round rewriting processes. At the prompt level, OPRO \cite{yang2023opro}, APE \cite{zhou2023ape}, DSPy \cite{khattab2024dspy}, EvoPrompt \cite{guo2024evoprompt}, PromptBreeder \cite{fernando2024promptbreeder}, and GEPA \cite{agrawal2026gepa} demonstrate that LLMs can iteratively rewrite prompts under score or reflection feedback; STOP \cite{zelikman2024stop}, Trace AutoDiff \cite{cheng2024trace}, and MARS \cite{zhang2026mars} extend the optimization scope to recursive self-optimization, execution trace feedback, and multi-agent collaboration. At coarser granularity, ReCreate \cite{hao2026recreate}, ReasoningBank \cite{ouyang2026reasoningbank}, DeltaEvolve \cite{jiang2026deltaevolve}, SE-Agent \cite{lin2025seagent}, AgentEvolver \cite{zhai2025agentevolver}, and EvolveR \cite{wu2026evolver} extend optimization targets to agent scaffolds, reasoning strategies, or experience lifecycles. APO \cite{pryzant2023apo} and TextGrad \cite{yuksekgonul2025textgrad} use natural language criticisms generated by LLMs as ``textual gradients,'' generalizing to arbitrary text variable optimization. The shared limitation of these methods is that accumulation states either remain at the version level (e.g., prompt iteration) or serve current-step edit selection (e.g., ranking and validation in online methods); no existing method maintains per-op stateful evidence across batches, so none can leverage reverse signals from subsequent batches to correct earlier judgments. MAA borrows the idea of ``organizing edits as an optimization process'' but achieves cross-batch evidence accumulation and correction through per-op EMA.

\subsection{Reinforcement Learning-Based Methods}

Unlike the non-parametric routes above, RL achieves self-evolution through gradient updates to model weights, which is orthogonal to MAA along the modification-object axis. PPO \cite{schulman2017ppo}, GRPO \cite{shao2024grpo}, RAGEN \cite{wang2025ragen}, and DPO \cite{rafailov2023dpo} are representative methods. RL requires GPU training and carries the risk of catastrophic forgetting; MAA only modifies context-injected content, requires no training, and leaves untouched memory items unchanged. The two can be deployed in combination in production.

\section{Method}

\begin{figure}[t]
\centering
\includegraphics[width=\linewidth]{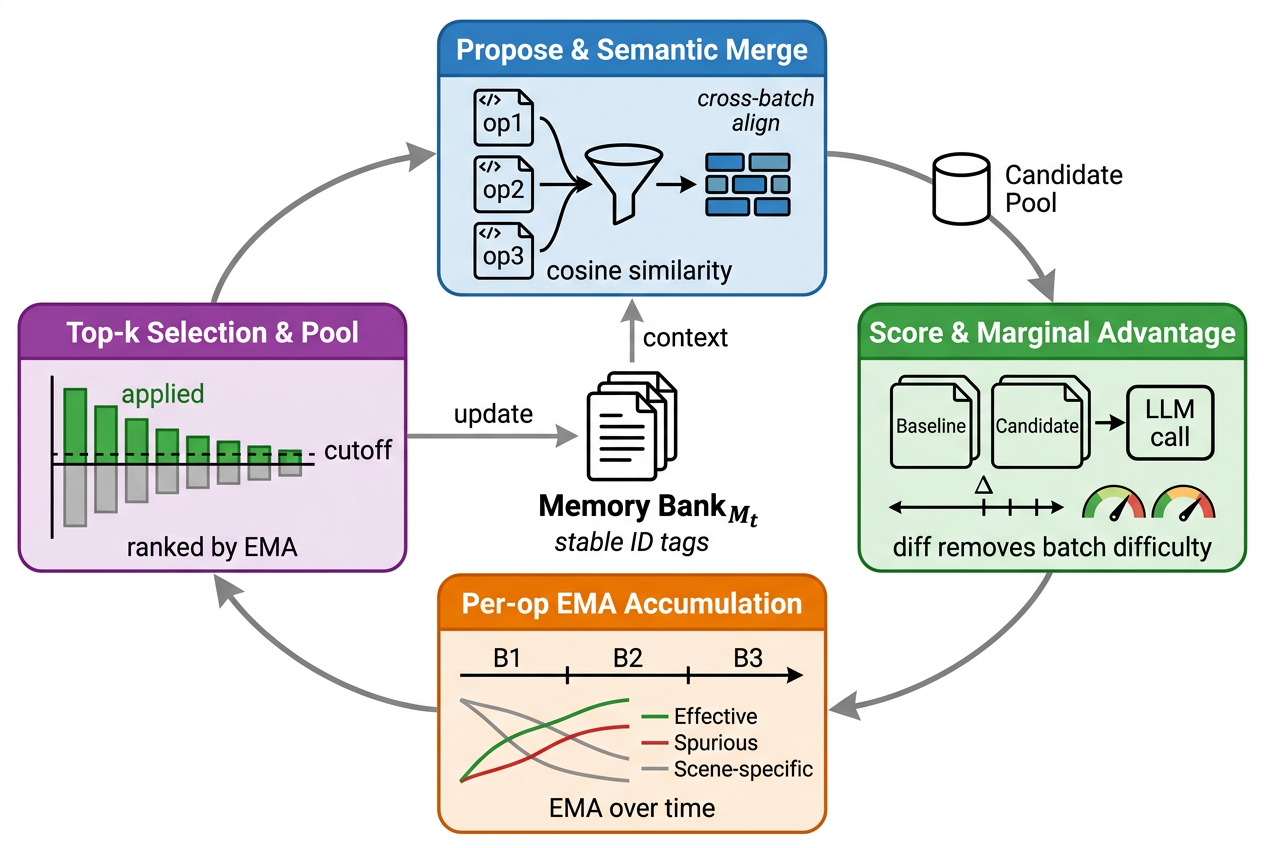}
\caption{MAA method overview. Propose channel generates candidate ops; Score channel constructs differential $\delta$; per-op EMA accumulates cross-batch evidence with top-$k$ selection.}
\label{fig:method_overview}
\end{figure}

This section describes MAA (Marginal Advantage Accumulation for Memory-Driven Agent Self-Evolution). Given a frozen base LLM $\phi$ and a fixed agent scaffold, the optimization objective is:
\begin{equation}
\max_{M \in \mathcal{M}} \; J(M) \triangleq \mathbb{E}_{q \sim \mathcal{D}}[\text{Eval}(\text{Agent}(q;\, \phi,\, M))]
\end{equation}
where $M$ is the only component in the context that can be intervened by the optimization process. Since trace collections far exceed a single context window, optimization proceeds in units of batch $B_t$. The two structural requirements established above (alignability and comparability) guide the designs in the following sections: \S3.1 defines the addressable memory bank and operation units, \S3.2 constructs marginal advantage signals, \S3.3 presents the per-op cross-batch accumulation mechanism, candidate pool management, and update budgets. The complete method flow is shown in Figure~\ref{fig:method_overview}.

\subsection{Addressable Memory Bank and Operation Units}

\textbf{Addressable memory bank.} At step $t$, when receiving task input $q$, the prompt constructed by the agent is $\text{Prompt}_t(q) = P_{\text{sys}} \,\Vert\, \text{Select}(M_t, q) \,\Vert\, q$. We abstract $M_t$ as an addressable set of text blocks:
\begin{equation}
M_t = \{m_1, m_2, \dots, m_n\}, \quad m_i = (\text{id}_i, c_i)
\end{equation}
where $\text{id}_i$ is assigned at creation, remains unchanged throughout the optimization cycle, and is not reused. Modifications to $M$ are implemented through two types of ops: $\text{add}$ (inserting new items) and $\text{modify}$ (rewriting or deleting existing items, with empty content for deletion). Each op is identified by a tuple $(\text{type},\; \text{anchor})$ plus a content vector, where anchor is the target\_id for modify or the position for add (only $\text{after:<id>}$ is supported).

\textbf{Semantic identity merging.} LLMs may propose semantically equivalent operations with different wordings across different batches. To avoid identity fragmentation, we use a lightweight embedding model (Qwen3-Embedding-4B) to compute semantic vectors for each op, merging new candidates into existing accumulation units only when type and anchor are identical and cosine similarity $\geq \tau$. Default $\tau = 0.85$. The choice of $\tau$ involves a precision--recall tradeoff: too high causes semantically equivalent ops to be scattered into multiple accumulation units, each unit's $t_k$ grows slowly, and top-$k$ truncation may push effective ops out of the candidate pool; too low causes ops with different intents to be incorrectly merged, mixing positive and negative signals and polluting the direction estimation of accumulation quantities. Within a reasonable range (e.g., $[0.80, 0.90]$), $\tau$ mainly affects evidence accumulation efficiency and ranking convergence speed, with limited impact on final performance (\S4.6.2 and Appendix B.5).

\textbf{Candidate operation generation.} At each step, the Propose channel receives $(M_t, B_t, \text{traces}_t)$ and outputs a set of add/modify candidate operations, which enter the candidate pool after semantic deduplication. The Propose stage does not generate Scores, preventing explanatory information from the candidate generation stage from leaking into the scoring stage and avoiding confirmation bias \cite{wang2024unfair} caused by the LLM's own generated explanations.

\subsection{Marginal Advantage}

\subsubsection{Differential Construction}

Absolute scoring $u(M, B)$ mixes memory quality, batch difficulty, and scoring noise; direct accumulation easily mistakes batch difficulty for op quality. Through same-batch baseline differencing, judgments are transformed into local comparisons, alleviating distribution differences between batches and making signals satisfy the comparability requirement. When $u$ exhibits interaction effects with batch difficulty (e.g., score compression in difficult batches, score inflation in easy batches), differencing still carries residual amplitude noise, which is smoothed and absorbed by subsequent cross-batch accumulation. The resulting differential signal $\delta_i$ is the marginal advantage of op $i$.

The Score channel estimates expected utility $u(M,B) \in [0,100]$: the likelihood of the agent achieving high-quality results on batch $B$ given memory $M$. This is an LLM-based proxy score rather than a true task score. Integer percentages are used to reduce the probability of $\delta = 0$ after differencing. The Score channel randomly shuffles the baseline $M_t$ and all candidate states $\{M_t^{(i)}\}$ within the same prompt for side-by-side valuation, and the optimizer performs outer-layer differencing to obtain marginal advantages:
\begin{equation}
\delta_i = u(M_t^{(i)}, B_t) - u(M_t, B_t) \in [-100, 100]
\end{equation}
$\delta_i$ takes the current baseline as reference, with sign determined by the differencing direction: candidate $u$ higher than baseline yields $\delta_i > 0$, and vice versa. When the number of candidates exceeds the grouping threshold, grouped batch valuation is adopted (evaluation details in Appendix A).

\subsubsection{Directional Alignment Assumption}

$\delta_i$ is not an unbiased gain estimate but a directional signal relative to the current baseline. $M_t^{(i)}$ is a hypothetical state not yet written to the memory bank; MAA relies on LLM proxy valuation on existing traces to avoid costly environment rollouts. The effectiveness of this signal depends on two properties.

\textbf{Internal signal quality.} The numerical value of $\delta$ is affected by absolute value drift in LLM scoring (systematic offsets from the same prompt under different context lengths or candidate orders) and inter-group scale drift introduced by grouped valuation, making single-step amplitudes not fully reliable. MAA eliminates position bias by randomly shuffling candidate order during side-by-side valuation, and mitigates inter-group drift by controlling group size. Although amplitudes are noisy, the sign of $\delta$ remains highly stable under perturbations, which is the prerequisite for the accumulation mechanism to function.

\textbf{External directional validity.} Let $\Delta\text{Eval}_i = \text{Eval}(M_i, B) - \text{Eval}(M, B)$ denote the true marginal gain of applying op $i$; the core condition supporting accumulation is directional alignment:
\begin{equation}
P\big(\text{sign}(\delta_i) = \text{sign}(\Delta\text{Eval}_i)\big) \geq 1/2 + \alpha, \quad \alpha > 0
\end{equation}
A smaller $\alpha$ requires more accumulation steps to produce stable rankings; if $\alpha \leq 0$, the Score channel cannot provide an effective optimization direction. When $\alpha > 0$, the sign of the EMA accumulation converges exponentially to the true direction as steps increase, consistent with classical analysis of sign-based stochastic optimization \cite{bernstein2018signsgd}. If $\alpha \leq 0$ in specific scenarios, EMA produces no systematic offset under zero-mean noise, and system performance degrades to no worse than Reactive Update. This means even if proxy scoring fails entirely, the accumulation mechanism will not introduce performance worse than the no-accumulation baseline. In theory, $u$ may have systematic directional bias toward certain types of ops, or $\alpha$ may be unevenly distributed across different operation types; such risks are common limitations of LLM-as-judge methods, which we discuss further in \S5 Limitations. Mechanism diagnosis experiments (\S4.5) sequentially verify internal signal quality through $\delta$-signal robustness and Sign Consistency, and verify whether directional alignment holds through Sign Accuracy.

\subsection{Cross-Batch Accumulation}

\subsubsection{Accumulation Operator and EMA Instance}

Cross-batch accumulation aggregates multiple marginal advantages $\delta_{k,t}$ of the same op across multiple batches (the differential signal of op $k$ at step $t$) into a stable accumulation quantity, allowing positive and negative evidence to cancel and consistent directions to be amplified. We choose exponential moving average (EMA) as the default accumulation operator, with the recursive form:
\begin{equation}
m_{k,t} = \beta \, m_{k,t-1} + (1 - \beta) \, \delta_{k,t}, \qquad \hat{m}_{k,t} = \frac{m_{k,t}}{1 - \beta^{t_k}}
\end{equation}
where $t_k$ is the cumulative number of updates op $k$ has participated in, and the denominator is the bias correction term. When $\delta_{k,\tau}$ signs are consistent across different batches, the magnitude of the accumulation quantity grows with steps; when signs alternate, positive--negative cancellation causes the accumulation quantity to trend toward zero. EMA simultaneously smooths residual amplitude noise from differencing, and through exponential weighting causes recent consistent-direction evidence to quickly dominate the accumulation quantity, accelerating ranking convergence.

The choice of EMA is based on its fit with the constraints of this design layer: magnitude retains the degree distinction between ``significant improvement'' and ``slight improvement,'' directly benefiting subsequent ranking; exponential decay matches the non-stationarity of memory states---the reference value of distant $\delta$ measured under old $M$ diminishes for current $M$. Running mean weights all historical $\delta$ equally, unable to adapt to signal distribution drift after memory updates; trimmed mean and median, although robust to outliers, discard amplitude information and incur computational overhead that grows with window size. EMA simultaneously achieves non-stationarity adaptation and amplitude retention with $O(1)$ recursion. $\beta = 0.9$ corresponds to an effective window of about 10 steps; budget decay, EMA exponential discounting, and the candidate pool maximum-age mechanism jointly constrain the drift speed of memory states, ensuring signals within the accumulation window correspond to recent memory states (specific parameters and numerical estimates in Appendix B).

\subsubsection{Candidate Pool and Update Budget}

The accumulation quantities $\hat{m}_{k,t}$ provide a unified ranking basis for candidate pool management and update decisions. Each op serves as an independent accumulation unit, with cross-batch identity consistency guaranteed by the semantic identity merging mechanism in \S3.1; all operations in the candidate pool are applied in-place on the current $M_t$ at each step and rescored, producing that step's $\delta_{k,t}$ and updating EMA.

The candidate pool $\mathcal{P}_t$ is preserved across steps. At each step, new candidates $\mathcal{C}_t^{\text{new}}$ are first merged with existing ops in the pool, then elimination is performed: candidates with $\hat{m}_k < m_{\text{floor}}$ are discarded; when pool size exceeds the upper limit, truncation is performed in descending order of $\hat{m}_k$; candidates surviving beyond $\text{max\_age}$ steps without being selected are eliminated. Default $m_{\text{floor}} = -50$, pool upper limit is 20, $\text{max\_age} = 10$.

Discrete memory items cannot undergo continuous small modifications, so top-$k$ is used as the per-step update count upper limit:
\begin{equation}
k_t = \min\!\Big(k_{\max},\; \max\!\big(k_{\min},\; \lfloor r_t \cdot |M_t|\rfloor\big)\Big), \quad r_t = r_{\max} - (r_{\max} - r_{\min})\cdot \frac{t}{T}
\end{equation}
$r_t$ linearly decays from $r_{\max}$ to $r_{\min}$, making exploration more thorough in early optimization and updates more conservative later. Default $r_{\max}=0.4,\, r_{\min}=0.1,\, k_{\min}=1,\, k_{\max}=8$. All operations are uniformly ranked by $\hat{m}_{k,t}$, and only the top-$k_t$ operations with $\hat{m}_{k,t} > 0$ are applied.

Best-checkpoint selection is performed at the end of each epoch based on $\text{Eval}$ scores on $\mathcal{D}_{\text{val}}$, ensuring return of the memory snapshot with the best validation set performance during training. Complete addressing rules, bias correction details, and algorithm pseudocode are provided in Appendix A.

\section{Experiments}

\subsection{Research Questions}

Experiments in this section revolve around four research questions.

\begin{itemize}
\item \textbf{RQ1: End-to-end effectiveness.} Does the complete MAA bring stable task benefits compared to the frozen (no memory) baseline, single-shot distillation, reactive update, offline distillation method Trace2Skill, and online method SkillOpt, under the same data split, same outer validation protocol, and similar inference budget?
\item \textbf{RQ2: Layer-by-layer ablation of cross-batch accumulation signals.} Through stepwise ablation from Reactive Update to Abs-score EMA, Counting-$\delta$ EMA, and finally Continuous-$\delta$ EMA, we sequentially verify the necessity of cross-batch accumulation, the irreplaceability of differential construction, and the additional gains of continuous amplitude for ranking.
\item \textbf{RQ3: Sign mechanism diagnosis.} Is the direction of differential $\delta$ highly consistent under perturbation despite limited robustness of differential amplitude $\delta$, and is the alignment with the true rollout gain direction significantly above chance?
\item \textbf{RQ4: Diagnostic analysis.} Does reactive update degenerate in long-term training due to accumulating harmful ops? Is MAA's evidence cancellation mechanism actually functioning?
\end{itemize}

\subsection{Experimental Setup}

\textbf{Datasets.} We evaluate on four benchmarks covering different interaction complexities (Table~\ref{tab:datasets}):

\begin{table}[htbp]
\centering
\caption{Dataset overview.}
\label{tab:datasets}
\begin{tabular}{@{}lllp{0.32\linewidth}@{}}
\toprule
Dataset & Type & Complexity & Capability Dimensions \\
\midrule
\textbf{ScienceAgentBench} \cite{chen2025scienceagentbench} & Scientific agent & High & Multi-step code generation, data analysis \\
\textbf{ALFWorld} \cite{shridhar2021alfworld} & Embodied agent & Medium-High & Multi-step decision-making, 6 task types \\
\textbf{HotpotQA} \cite{yang2018hotpotqa} & Multi-hop QA & Low & Multi-hop factual retrieval \\
\textbf{SpreadsheetBench} \cite{ma2024spreadsheetbench} & Spreadsheet & Medium & Formula reasoning, data manipulation \\
\bottomrule
\end{tabular}
\end{table}

All datasets use 500 training trajectories; ScienceAgentBench, positioned as a verification benchmark, generates training data independently in the corresponding environment, with all 88 original tasks used for testing. Dataset splitting details and usage protocols are provided in Appendix B.1.

\textbf{Comparison methods.} The RQ1 main experiment includes the following baselines (Table~\ref{tab:baselines}):

\begin{table}[htbp]
\centering
\caption{Comparison methods and their alignment with MAA.}
\label{tab:baselines}
\begin{tabular}{@{}lll@{}}
\toprule
Method Category & Representative Method & Alignment \\
\midrule
\textbf{Frozen (No-Memory)} & Base prompt + tool hints only & No self-evolution, lower-bound reference \\
\textbf{Single-shot distillation} & ExpeL / Trace2Skill-style & One-time extraction, then frozen \\
\textbf{Reactive update} & MAA's no-accumulation ablation & Same channels, no EMA or pool \\
\textbf{Offline hier.\ distillation} & Trace2Skill \cite{ni2026trace2skill} & Offline hierarchical induction \\
\textbf{Online skill optimization} & SkillOpt \cite{yang2026skillopt} & Online rollout, skill-level optimization \\
\textbf{MAA (Ours)} & Full system & Differencing + per-op EMA + top-$k$ \\
\bottomrule
\end{tabular}
\end{table}

SkillOpt \cite{yang2026skillopt} and Trace2Skill \cite{ni2026trace2skill} are both reproduced using official implementations with the original papers' default hyperparameters unchanged; reproduction details are provided in Appendix B.2.

\textbf{Target models.} We validate generalizability on four target models: Qwen3.7-Max (flagship strong model, all experiments), Qwen3.6-Flash (lightweight weak model, RQ1), DeepSeek-V4-Flash (cross-family lightweight model, RQ1), and GPT-5.4 (cross-family strong model, RQ1 only), covering the full ``strong/weak $\times$ same-family/cross-family'' 2$\times$2 design space. The Score channel uniformly uses Qwen3.7-Max.

\textbf{Experiment configuration overview.} RQ1 main results use 5 independent random seeds per configuration, reporting mean $\pm$ standard deviation. Specific sample sizes, calibration set settings, and evaluation metrics for each RQ are provided in Appendix B.4--B.5.

\subsection{End-to-End Main Results}

\subsubsection{Main Experimental Results}

Table~\ref{tab:end2end} reports test set scores (5 seeds, mean $\pm$ std) of each method across four datasets $\times$ four target models.

\begin{table}[htbp]
\centering
\caption{End-to-end task performance (test set scores, 5 seeds mean $\pm$ std). ScienceAgentBench reports code pass rate (\%), ALFWorld reports task success rate (\%), SpreadsheetBench reports accuracy (\%), HotpotQA reports Exact Match (\%). Bold indicates best in each column.}
\label{tab:end2end}
\small
\begin{tabular}{@{}lcccc@{}}
\toprule
Method & ScienceAgentBench & ALFWorld & SpreadsheetBench & HotpotQA \\
\midrule
& \textbf{Qwen3.7-Max} & & & \\
Frozen (no memory) & 22.1 $\pm$ 2.1 & 81.0 $\pm$ 1.8 & 41.8 $\pm$ 1.6 & 71.1 $\pm$ 0.9 \\
Single-shot (ExpeL-style) & 25.8 $\pm$ 2.9 & 83.5 $\pm$ 2.0 & 47.3 $\pm$ 1.8 & 73.6 $\pm$ 1.2 \\
Reactive Update & 28.4 $\pm$ 3.3 & 85.2 $\pm$ 1.7 & 51.4 $\pm$ 2.0 & 74.5 $\pm$ 1.0 \\
Trace2Skill & 27.2 $\pm$ 2.8 & 85.8 $\pm$ 1.9 & 52.6 $\pm$ 1.7 & 75.1 $\pm$ 1.0 \\
SkillOpt \cite{yang2026skillopt} & 30.3 $\pm$ 2.6 & 88.1 $\pm$ 1.5 & 56.2 $\pm$ 1.5 & 76.9 $\pm$ 0.7 \\
\textbf{MAA (Ours)} & \textbf{30.7 $\pm$ 2.4} & \textbf{89.4 $\pm$ 1.6} & \textbf{58.5 $\pm$ 1.6} & \textbf{77.2 $\pm$ 1.3} \\
\midrule
& \textbf{Qwen3.6-Flash} & & & \\
Frozen (no memory) & 11.3 $\pm$ 2.7 & 76.8 $\pm$ 2.4 & 32.8 $\pm$ 2.2 & 65.4 $\pm$ 1.5 \\
Single-shot (ExpeL-style) & 15.7 $\pm$ 2.8 & 79.5 $\pm$ 2.1 & 37.4 $\pm$ 1.9 & 68.5 $\pm$ 1.7 \\
Reactive Update & 18.3 $\pm$ 3.4 & 81.6 $\pm$ 2.3 & 41.6 $\pm$ 2.4 & 68.8 $\pm$ 1.4 \\
Trace2Skill & 17.5 $\pm$ 3.0 & 81.2 $\pm$ 2.2 & 42.8 $\pm$ 2.0 & 70.2 $\pm$ 1.5 \\
SkillOpt \cite{yang2026skillopt} & \textbf{22.6 $\pm$ 2.9} & 84.2 $\pm$ 2.2 & 46.8 $\pm$ 1.8 & 71.8 $\pm$ 1.3 \\
\textbf{MAA (Ours)} & 20.8 $\pm$ 2.7 & \textbf{86.2 $\pm$ 2.0} & \textbf{48.5 $\pm$ 2.1} & \textbf{72.6 $\pm$ 1.6} \\
\midrule
& \textbf{GPT-5.4} & & & \\
Frozen (no memory) & 22.7 $\pm$ 2.6 & 80.5 $\pm$ 1.8 & 42.6 $\pm$ 1.7 & 70.8 $\pm$ 1.1 \\
Single-shot (ExpeL-style) & 26.1 $\pm$ 2.5 & 83.0 $\pm$ 1.7 & 47.8 $\pm$ 1.6 & 73.7 $\pm$ 1.2 \\
Reactive Update & 29.2 $\pm$ 2.8 & 84.8 $\pm$ 1.6 & 51.9 $\pm$ 1.3 & 74.1 $\pm$ 1.0 \\
Trace2Skill & 28.6 $\pm$ 2.6 & 85.3 $\pm$ 1.5 & 52.7 $\pm$ 1.5 & 74.6 $\pm$ 1.5 \\
SkillOpt \cite{yang2026skillopt} & 31.9 $\pm$ 2.4 & 87.5 $\pm$ 1.2 & 56.8 $\pm$ 1.4 & 77.2 $\pm$ 0.9 \\
\textbf{MAA (Ours)} & \textbf{32.8 $\pm$ 2.3} & \textbf{88.4 $\pm$ 1.5} & \textbf{58.5 $\pm$ 1.6} & \textbf{78.0 $\pm$ 1.4} \\
\midrule
& \textbf{DeepSeek-V4-Flash} & & & \\
Frozen (no memory) & 14.8 $\pm$ 3.5 & 74.3 $\pm$ 2.3 & 35.6 $\pm$ 2.1 & 62.8 $\pm$ 1.3 \\
Single-shot (ExpeL-style) & 19.8 $\pm$ 2.7 & 77.8 $\pm$ 1.9 & 40.8 $\pm$ 1.7 & 65.7 $\pm$ 1.4 \\
Reactive Update & 19.5 $\pm$ 3.3 & 79.4 $\pm$ 2.5 & 44.5 $\pm$ 2.3 & 66.1 $\pm$ 1.0 \\
Trace2Skill & 20.4 $\pm$ 2.9 & 80.5 $\pm$ 2.1 & 45.2 $\pm$ 2.0 & 66.9 $\pm$ 1.2 \\
SkillOpt \cite{yang2026skillopt} & \textbf{23.4 $\pm$ 3.1} & 83.2 $\pm$ 1.8 & 50.6 $\pm$ 1.8 & 68.3 $\pm$ 1.2 \\
\textbf{MAA (Ours)} & 21.7 $\pm$ 2.8 & \textbf{84.9 $\pm$ 2.1} & \textbf{51.2 $\pm$ 1.9} & \textbf{68.9 $\pm$ 1.2} \\
\bottomrule
\end{tabular}
\end{table}

MAA achieves the best results in 14 out of 16 settings. On the two strong models Qwen3.7-Max and GPT-5.4, MAA outperforms SkillOpt on all four datasets (GPT-5.4: ScienceAgentBench +0.9 pp, ALFWorld +0.9 pp, SpreadsheetBench +1.7 pp, HotpotQA +0.8 pp), showing that offline proxy scoring suffices to guide optimization direction on complex tasks, and the advantage observed on the cross-family strong model (GPT-5.4) further supports MAA's generalizability. On Qwen3.6-Flash and DeepSeek-V4-Flash, the pattern diverges: MAA still achieves the best on ALFWorld, SpreadsheetBench, and HotpotQA, but is surpassed by SkillOpt on ScienceAgentBench (Flash +1.8 pp, DeepSeek +1.7 pp). MAA outperforms the offline architecture Trace2Skill in all 16 settings, reflecting the advantage of cross-batch evidence accumulation over one-shot hierarchical induction.

The gap between online and offline methods diverges with model capability. On strong models (Qwen3.7-Max, GPT-5.4), MAA surpasses SkillOpt on ScienceAgentBench because strong models produce higher-quality traces, making the proxy scoring directional signal $\alpha$ large enough for EMA to converge effectively. Weak models (Flash / DeepSeek) are surpassed by SkillOpt on this dataset because their failure modes in real environments are more diverse, offline traces are noisier making $\alpha$ smaller and accumulation convergence slower; whereas online rollout provides direct observation of true execution consequences, not relying on proxy scoring direction accuracy, enabling more efficient correction. When the base model is weak and the task is difficult, online real-time feedback is more efficient at correction than offline proxy accumulation.

MAA's improvement over Reactive Update increases with task complexity. On Qwen3.7-Max: SpreadsheetBench +7.1 pp, ALFWorld +4.2 pp, ScienceAgentBench +2.3 pp, HotpotQA +2.7 pp; GPT-5.4 shows the same trend (SpreadsheetBench +6.6 pp, ALFWorld +3.6 pp, ScienceAgentBench +3.6 pp, HotpotQA +3.9 pp); Qwen3.6-Flash shows the same trend. The more complex the task and the greater the inter-batch distribution differences, the more valuable cross-batch evidence accumulation becomes. On ScienceAgentBench, MAA's improvement over Reactive is similar to HotpotQA on strong models, yet MAA ultimately surpasses SkillOpt, suggesting that the accumulation mechanism still converges to effective rankings on complex tasks under strong models.

\subsubsection{Computational Resource Consumption Comparison}

MAA and the online method (SkillOpt) share the same Propose + Score overhead; the core difference is that the online method requires additional environment rollout at each step. Table~\ref{tab:resource} quantifies this difference.

\begin{table}[htbp]
\centering
\caption{Computational resource consumption comparison (ALFWorld, Qwen3.7-Max, single full training run).}
\label{tab:resource}
\begin{tabular}{@{}lll@{}}
\toprule
Metric & MAA (Ours) & SkillOpt \cite{yang2026skillopt} \\
\midrule
\textbf{Optimization-phase rollouts} & \textbf{0} & $\sim$12.8K \\
\textbf{Token consumption (M)} & $\sim$8.5 & $\sim$33.3 \\
\textbf{Estimated API cost (USD)} & \textbf{$\sim\$\text{30}$} & $\sim\$\text{120}$ \\
\textbf{Training time (min)} & $\sim$145 & $\sim$750--830 \\
\textbf{Test score (\%)} & \textbf{89.4} & 88.1 \\
\bottomrule
\end{tabular}
\end{table}

\emph{Only the optimization phase is counted, excluding Val/Test evaluation. API costs estimated based on Qwen3.7-Max public pricing.}

MAA achieves better test scores with approximately 1/4 the token consumption and 1/5 the training time. The offline architecture trades lower computational cost for dependence on existing traces.

\subsection{Design Choices for Accumulation Signals (RQ2)}

This section verifies the design choices for accumulation signals: the necessity of differential construction and the additional benefit of continuous amplitude. Abs-score EMA appears only as a targeted ablation in this section.

\subsubsection{Experimental Design}

We run layer-by-layer ablation on all four tasks using Qwen3.7-Max (dataset descriptions in \S4.2). The four configurations share the same Propose / Score channels, candidate pool size, semantic identity merging threshold ($\tau=0.85$), top-$k$ selection, best-checkpoint, training data split, and training step budget, differing only in the accumulation signal (Table~\ref{tab:ablation_design}):

\begin{table}[htbp]
\centering
\caption{Ablation study design for accumulation signals.}
\label{tab:ablation_design}
\begin{tabular}{@{}lllll@{}}
\toprule
Variant & Cross-batch & Differencing & Signal & Verified \\
\midrule
Reactive Update & $\times$ & --- & Current batch feedback & No-accum. reference \\
Abs-score EMA & $\checkmark$ & $\times$ & $u(M_i,B)$ & Necessity of diff. \\
Counting-$\delta$ EMA & $\checkmark$ & $\checkmark$ & $\text{sign}(\delta)$ & Continuous benefit \\
Continuous-$\delta$ EMA & $\checkmark$ & $\checkmark$ & Continuous $\delta$ & Full system \\
\bottomrule
\end{tabular}
\end{table}

Abs-score EMA uses $m_{i,t}=\beta m_{i,t-1}+(1-\beta)u(M_i,B_t)$, Counting-$\delta$ EMA uses $s_{i,t}=\text{sign}(\delta_{i,t})\in\{-1,0,+1\}$, and Continuous-$\delta$ EMA uses the raw differential $\delta_{i,t}=u(M_i,B_t)-u(M,B_t)$. All EMA configurations use $\beta=0.9$, with each configuration run on 5 random seeds.

\subsubsection{Experimental Results and Analysis}

\begin{table}[htbp]
\centering
\caption{Layer-by-layer ablation of accumulation signals (Qwen3.7-Max, 5 seeds).}
\label{tab:ablation_results}
\begin{tabular}{@{}lcccc@{}}
\toprule
Variant & SAB (\%) & ALF (\%) & SS (\%) & HQA (\%) \\
\midrule
Reactive Update & 28.4 $\pm$ 2.3 & 85.2 $\pm$ 1.7 & 51.4 $\pm$ 2.0 & 74.5 $\pm$ 1.0 \\
Abs-score EMA & 28.9 $\pm$ 2.1 & 86.3 $\pm$ 1.8 & 53.8 $\pm$ 2.1 & 75.1 $\pm$ 1.1 \\
Counting-$\delta$ EMA & 29.3 $\pm$ 1.8 & 87.8 $\pm$ 1.9 & 57.2 $\pm$ 2.1 & 76.1 $\pm$ 1.0 \\
\textbf{Continuous-$\delta$ EMA (MAA)} & \textbf{30.7 $\pm$ 2.4} & \textbf{89.4 $\pm$ 1.6} & \textbf{58.5 $\pm$ 1.6} & \textbf{77.2 $\pm$ 1.3} \\
\bottomrule
\end{tabular}
\end{table}

Table~\ref{tab:ablation_results} reports the ablation results. \textbf{Differential construction is the primary source of accumulation gains.} Although Abs-score EMA outperforms Reactive Update (+0.5--2.4 pp), it remains significantly below the full MAA (gap of 1.8--4.7 pp). Raw $u$ mixes memory quality, batch difficulty, and scoring noise; direct accumulation easily mistakes batch difficulty for op quality. Counting-$\delta$ EMA clearly outperforms Abs-score EMA on all tasks, demonstrating that baseline differencing provides signed directional evidence, enabling cross-batch positive--negative cancellation.

\textbf{Continuous amplitude provides fine-grained ranking gains.} Continuous-$\delta$ EMA still shows +1.1--1.6 pp advantage over Counting-$\delta$ EMA, but this gap is smaller than the gain from differential construction, indicating that MAA's primary benefit comes from signed marginal evidence, with continuous amplitude mainly supplementing ranking precision.

\textbf{Improvement magnitude correlates with proxy signal quality.} Total improvement from Reactive$\to$MAA: SpreadsheetBench (+7.1 pp) $>$ ALFWorld (+4.2 pp) $>$ HotpotQA (+2.7 pp) $>$ ScienceAgentBench (+2.3 pp). ScienceAgentBench shows the smallest improvement, consistent with its lowest Sign accuracy (61.5\%) and longer chains causing larger directional bias; the improvement differences among the other three tasks are simultaneously influenced by the magnitude of inter-batch distribution differences and the baseline's room for improvement, and are not determined by any single factor alone.

\subsection{Mechanism Diagnosis: Signal Robustness and Directional Alignment (RQ3)}

\S3.2.2 establishes that the effectiveness of differential signals depends on two properties: internal signal quality (stability of $\delta$ amplitude under perturbation) and external directional validity (directional alignment, i.e., the probability that $\text{sign}(\delta)$ aligns with the true gain direction $\geq 1/2 + \alpha$). This section sequentially verifies these two properties through three metrics. We apply prompt perturbation (candidate order shuffling + context length variants) to the same $(M_i, B_t)$ and rescore, measuring: $\delta$-signal robustness (Pearson $r$ between two $\delta$ values, verifying internal signal quality), Sign consistency (proportion of $\text{sign}(\delta)$ remaining unchanged under perturbation, verifying directional stability), and Sign accuracy (proportion of $\text{sign}(\delta_i)$ aligning with true rollout gain direction, verifying directional alignment).

\begin{table}[htbp]
\centering
\caption{Differential signal robustness and direction diagnosis (Qwen3.7-Max, \%).}
\label{tab:signal_robustness}
\begin{tabular}{@{}lcccc@{}}
\toprule
Metric & SAB & ALF & SS & HQA \\
\midrule
$\delta$-signal robustness & 62.3 & 76.2 & 64.1 & 78.1 \\
Sign consistency & \textbf{91.2} & \textbf{95.1} & \textbf{88.7} & \textbf{93.6} \\
Sign accuracy & \textbf{61.5} & \textbf{72.1} & \textbf{69.3} & \textbf{73.8} \\
\bottomrule
\end{tabular}
\end{table}

Table~\ref{tab:signal_robustness} summarizes the results. Single-step numerical robustness of $\delta$ is limited (62.3\%--78.1\%), but Sign consistency exceeds 88.7\% on all datasets, indicating that $\delta$ signs are highly stable under perturbation. Sign accuracy corresponds to $\alpha \approx 0.11$--$0.24$, with directional alignment holding on all datasets and decreasing with task complexity. ScienceAgentBench's $\alpha \approx 0.115$ is close to the lower bound, requiring more accumulation steps to produce stable rankings; nevertheless, Table~\ref{tab:ablation_results} shows MAA still outperforms Reactive by +2.3 pp on this dataset, showing that even with the weakest directional signal, cross-batch accumulation can still extract effective optimization directions. The gain space is limited by the magnitude of $\alpha$, with correspondingly smaller improvement, and on weak models MAA is surpassed by the online method SkillOpt. Monte Carlo simulation further quantifies the relationship between convergence speed and $\alpha$. Under $\beta=0.9$ and the conservative symmetric $|\delta|=1$ assumption, $\alpha=0.183$--$0.238$ (corresponding to SpreadsheetBench, ALFWorld, HotpotQA) requires 7--14 updates for the EMA sign to stabilize to the true direction with $\geq$90\% probability; $\alpha=0.115$ (ScienceAgentBench) requires more than 10 updates under the symmetric assumption, but experiments confirm MAA still outperforms Reactive by +2.3 pp, suggesting that the asymmetry of the actual $\delta$ distribution compensates for the theoretical convergence gap.

\subsection{Design Layer Verification: Learning Curves and Evidence Trajectories}

\begin{figure}[t]
\centering
\includegraphics[width=\linewidth]{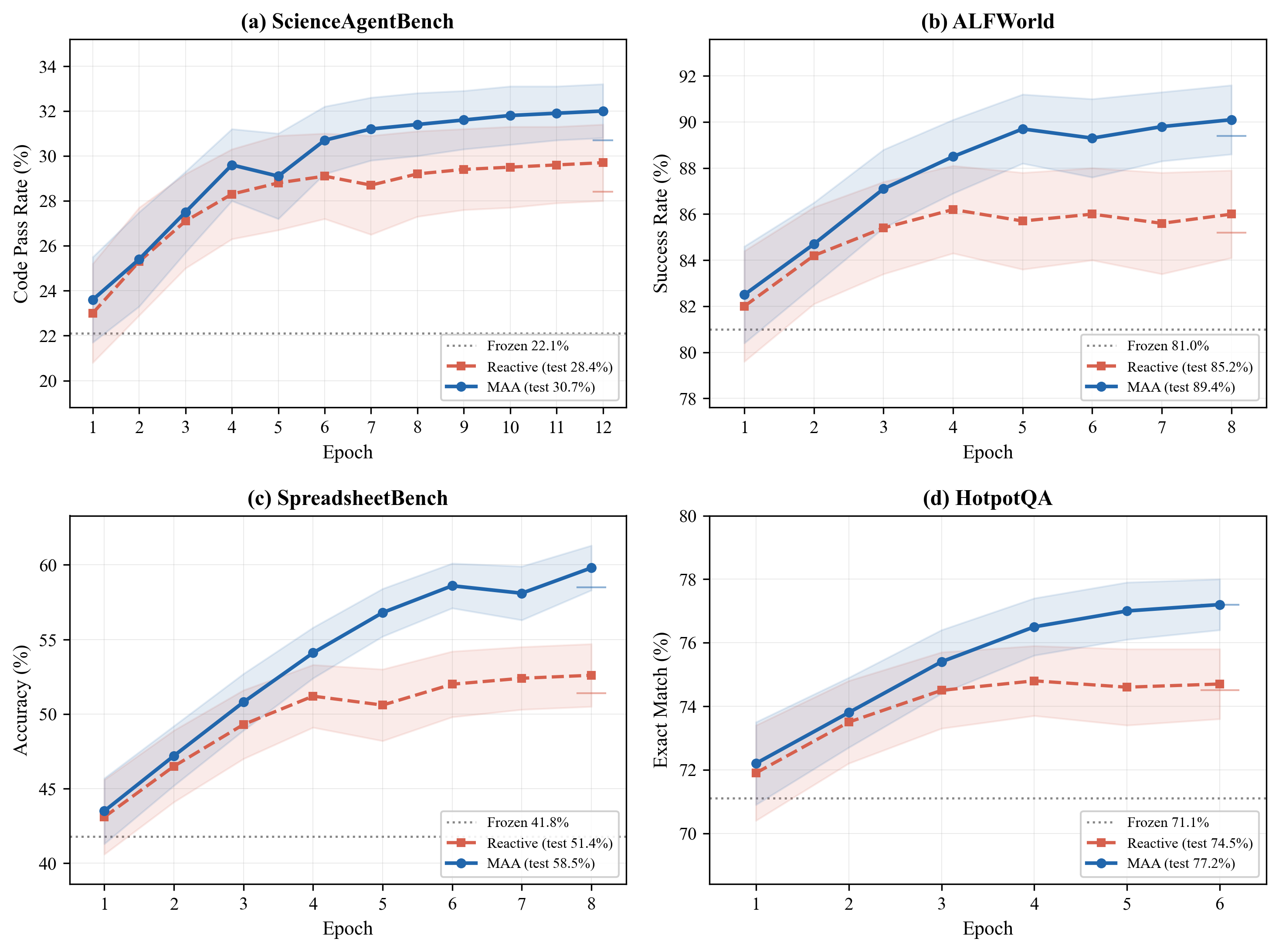}
\caption{Learning curves (Qwen3.7-Max, 5 seeds). Solid: MAA; dashed: Reactive; dotted: Frozen baseline. Shaded bands: $\pm$1 std.}
\label{fig:learning_curves}
\end{figure}

This experiment corresponds to RQ4, visualizing the actual operation of MAA's cross-batch evidence accumulation through learning curves and evidence trajectories (Figure~\ref{fig:learning_curves}).

\subsubsection{Learning Curve Comparison}

On all four datasets, MAA opens a gap with Reactive at epoch 3--4, but convergence speed diverges significantly with task complexity: HotpotQA (simple multi-hop QA) reaches test-level performance within 6 epochs; ALFWorld and SpreadsheetBench (medium complexity) require 8 epochs to converge; ScienceAgentBench (long-chain scientific discovery) requires about 12 epochs to stabilize. Convergence speed is consistent with task state space size, trajectory length, and $\delta$ signal-to-noise ratio. The ranking of MAA's gap relative to Reactive aligns with the ablation improvement in \S4.4 and the Sign accuracy in \S4.5, with training dynamics observations matching the preceding analysis.

\subsubsection{Evidence Trajectories and Case Study}

\begin{figure}[t]
\centering
\includegraphics[width=\linewidth]{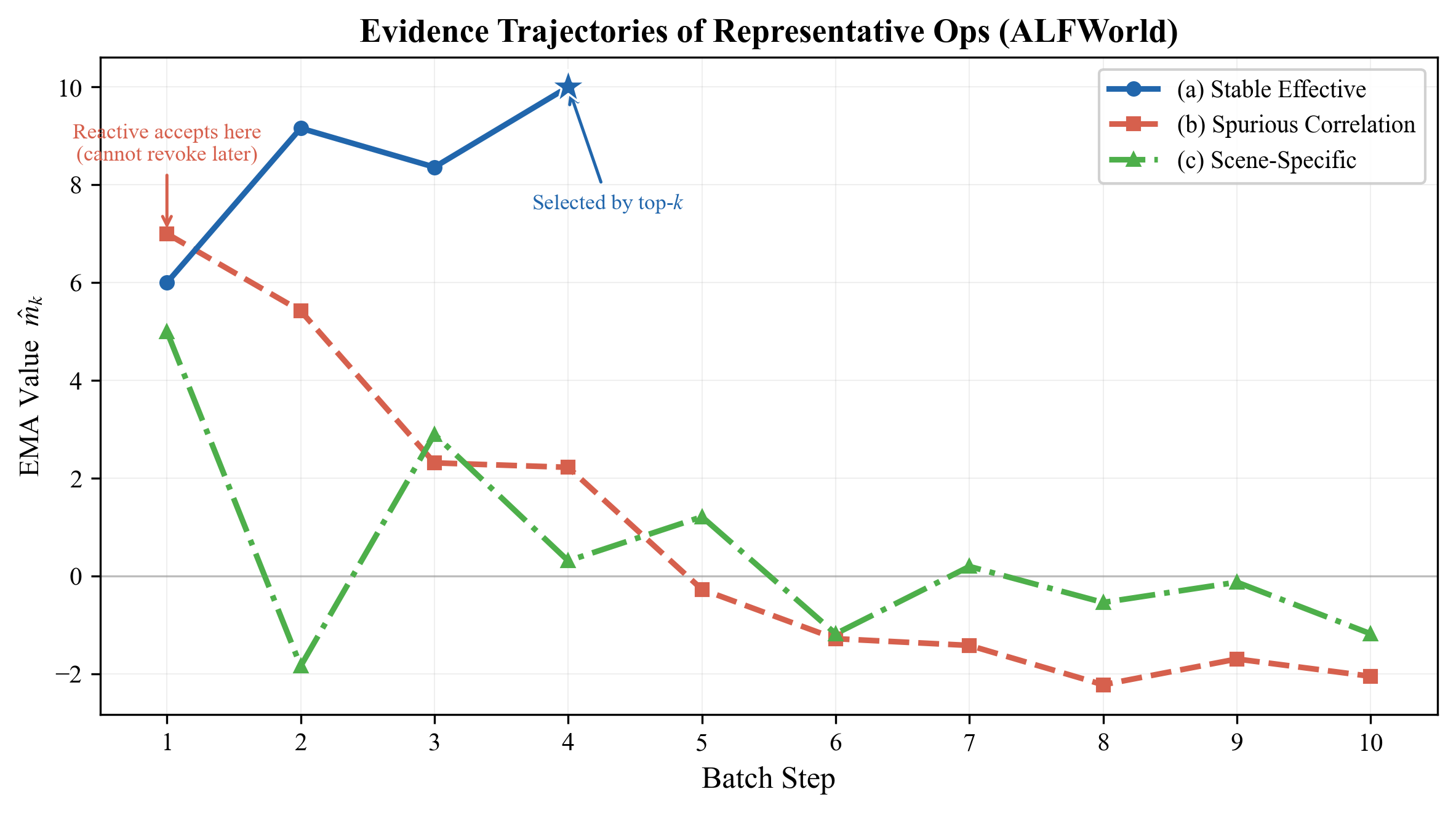}
\caption{EMA evidence trajectories of representative ops (ALFWorld). Blue: stably effective; red: spurious correlation; green: scene-specific.}
\label{fig:evidence_trajectories}
\end{figure}

Figure~\ref{fig:evidence_trajectories} shows the EMA value $\hat{m}_k$ trajectories of three representative ops across batch steps during ALFWorld training, corresponding to three typical patterns.

\begin{itemize}
\item \textbf{(a) Stable Effective} (``{[}modify \#7{]} If the target is not found in the current container after examination, immediately switch to the next most likely container.''): A general navigation strategy, $\delta$ consistently positive (+6, +12, +7, +14), bias-corrected EMA converges to $\approx$+10 within 4 steps, selected by top-$k$ and written to memory. When directions are consistent, EMA converges rapidly, as expected.
\item \textbf{(b) Spurious Correlation} (``{[}add after \#3{]} Heat tasks should check microwave before oven''): The first two batches happen to be applicable ($\delta$ = +7, +4), and reactive update accepts it at batch 1 with no way to retract. Subsequent $\delta$ mean leans negative ($-$3, +2, $-$8, $-$5, $-$2, $-$6, +1, $-$4), bias-corrected EMA gradually declines from +7.0 to $-$0.7, and is eliminated by max\_age after 10 steps. Differential positive--negative cancellation filters this operation, avoiding reactive update's irreversible error.
\item \textbf{(c) Scene-Specific} (``{[}modify \#12{]} In cleaning tasks, inspect the sink area for the cloth first''): $\delta$ alternates between positive and negative (+5, $-$8, +11, $-$6, +4, $-$10, +6, $-$4, +2, $-$7), bias-corrected EMA oscillates near zero (final $\approx$ $-$1.2), and is eliminated by max\_age after 10 steps. When directions are inconsistent, EMA produces no systematic offset, ensuring degradation safety.
\end{itemize}

\textbf{Semantic identity merging and $\tau$ selection.} Alignability requires the same operation to be trackable across batches. During ALFWorld training, at $\tau=0.85$, 15.7\% of new candidates were merged; at $\tau=0.80$ this rises to 19.2\% but incorrect merging occurs (same anchor, different intent); at $\tau=0.90$ it drops to 12.4\%, with some semantically equivalent ops not recognized due to wording differences. $\tau=0.85$ achieves a balance between incorrect merging and missed merging. A typical case is ``{[}modify \#7{]}'' in Figure~\ref{fig:evidence_trajectories}(a): 3 different wordings were merged into the same unit, and bias-corrected EMA converged to $\approx$+10 at step 4; without merging, this strategy would be lost due to identity fragmentation. The low merge rate of 15.7\% indicates the mechanism is precise, triggering only when semantics are truly equivalent. More merging cases (including correct rejections and missed merging scenarios) are detailed in Appendix B.5.

\section{Conclusion}

In batch-style trace distillation, feedback received by the same memory operation across different batches is often inconsistent, making it difficult to judge from single-step signals whether it is broadly effective or merely an accidental hit in specific batches. The more complex the task and the longer the trajectory, the more prominent this problem becomes.

MAA maintains a cross-batch accumulated signed evidence quantity for each operation. Differential construction transforms absolute scores into marginal advantages relative to the current baseline, making signals across different batches comparable; EMA temporally aggregates these marginal advantages with exponential weighting, amplifying directionally consistent signals and canceling directionally alternating ones. This mechanism does not rely on environment rollouts, using only LLM proxy scores as directional signals, completing operation-level screening and ranking under offline settings. From ablation to mechanism diagnosis to training dynamics, experiments consistently show that differencing and accumulation layers jointly form the performance source of MAA.

Experimental results show that MAA achieves the best result in 14 out of 16 settings across 4 datasets and 4 target models, with token consumption reduced by approximately 75\% and optimization time shortened from 12--14 hours to about 2.5 hours. In scenarios where forward evaluation is costly, the offline accumulation strategy achieves performance comparable to or even better than online methods with significantly lower resource overhead.

\subsection{Limitations}

\textbf{Offline architecture scenario positioning.} MAA replaces environment rollouts with LLM proxy scoring, making it suitable for scenarios with high forward evaluation costs (scientific code execution, real user interaction logs, API calls with side effects, etc.). When forward evaluation is sufficiently cheap and the environment is reproducible, directly using val scores as the optimization signal may be a simpler choice. Additionally, when memory undergoes substantial changes within a single step, the coverage of old traces for the updated state may decrease; in practice, this can be diagnosed by monitoring whether $\delta$ variance increases abnormally, and trace refresh can be triggered when necessary.

\textbf{Systematic bias of LLM judge.} The Score channel relies on LLM proxy valuation, which may be affected by position bias, length bias, and other factors. We mitigate such biases through random shuffling of candidate order and differential construction (side-by-side valuation of baseline and candidates within the same prompt); \S4.5 mechanism diagnosis shows $\delta$ signs are highly consistent (Sign consistency $\geq$ 88.7\%) and significantly aligned with the true gain direction (Sign accuracy $\geq$ 61.5\%). Residual biases are common limitations of LLM-as-judge methods, but have not substantially impacted MAA's effectiveness within the task range covered by current experiments.

\textbf{Experimental coverage scope.} Current experiments cover 4 benchmarks and 4 target models (Qwen3.7-Max, Qwen3.6-Flash, GPT-5.4, DeepSeek-V4-Flash), with core hyperparameters ($\beta=0.9$, $\tau=0.85$, $m_{\text{floor}}=-50$) determined through independent pilot tuning, with pilot splits not shared with formal experiments to avoid $\mathcal{D}_{\text{val}}$ contamination. Ablation experiments show that MAA's advantages over various variants are consistent across multiple datasets, supporting the robustness of the current configuration; broader cross-domain generalization verification is left for future work.

\subsection{Future Work}

We plan to verify whether the value of cross-batch accumulation is further amplified in longer trajectories and tool-call-intensive scenarios, and explore combined deployment of MAA as an offline signal source with online RL, leveraging the complementary advantages of both in signal quality and feedback timeliness. Additionally, adaptive mechanisms for the semantic identity merging threshold $\tau$ and the applicability of alternative accumulation operators (such as counting EMA) under different LLM backbones warrant further research. We will also conduct hyperparameter sensitivity analysis to further verify the method's robustness under different configurations.

\bibliographystyle{unsrtnat}
\bibliography{references}

\section*{Appendix A: Complete Algorithm Pseudocode}
\addcontentsline{toc}{section}{Appendix A: Complete Algorithm Pseudocode}

\textbf{Notation conventions.} $u(M,B) \in [0,100]$ is the Score channel's LLM proxy valuation (\S3.2), used only for internal differential signals during training; $\text{Eval}(M;\mathcal{D})$ is the outer-layer true task evaluation (running the agent on each item in $\mathcal{D}$ driven by $M$, computing pass rate with objective metrics). $\mathcal{D}_{\text{train}}$, $\mathcal{D}_{\text{val}}$, $\mathcal{D}_{\text{test}}$ are mutually disjoint; Propose / Score only touches $\mathcal{D}_{\text{train}}$; $\mathcal{D}_{\text{val}}$ calls $\text{Eval}$ only once at initialization and once at the end of each epoch; $\mathcal{D}_{\text{test}}$ is evaluated only once throughout. All baselines follow the same protocol.

\textbf{Default hyperparameters.}

\begin{table}[htbp]
\centering
\begin{tabular}{@{}lll@{}}
\toprule
Parameter & Value & Description \\
\midrule
$\beta$ & 0.9 & EMA decay coefficient, effective window $\sim$10 steps \\
$\tau$ & 0.85 & Cosine similarity threshold for semantic identity merging \\
$m_{\text{floor}}$ & $-$50 & Candidate pool advantage lower bound \\
$\text{pool\_size\_max}$ & 20 & Candidate pool size upper limit \\
$\text{max\_age}$ & 10 & Maximum global survival steps for unselected ops \\
$r_{\max}, r_{\min}$ & 0.4, 0.1 & Start/end values for top-$k$ ratio linear decay \\
$k_{\min}, k_{\max}$ & 1, 8 & Top-$k$ count lower and upper bounds \\
$g$ & Context constraint & Upper limit of states in a single Score call \\
\bottomrule
\end{tabular}
\end{table}

\textbf{Algorithm.}

\begin{verbatim}
Input: Initial memory M_0; pre-collected traces {tau_q : q in D_train};
       Datasets D_train, D_val, D_test (mutually disjoint)
Output: Eval(M*; D_test)

# ---- Initialization ----
M <- M_0;  P <- empty;  EMA <- {}
best_score <- Eval(M_0; D_val);  M* <- M_0
patience_counter <- 0;  global_step <- 0

# ---- Training main loop ----
for epoch in 1..max_epochs:
    for inner_step in 1..steps_per_epoch:
        global_step <- global_step + 1;  t <- global_step

        # 1. Sample batch
        B_t <- Sample(D_train, batch_size)
        traces_t <- Traces[B_t]

        # 2. Propose: Generate candidate ops (S3.1)
        C_new <- Propose(M, B_t, traces_t)

        # 3. Candidate pool maintenance
        # 3a) Semantic identity merging (S3.1)
        for op in C_new:
            e_new <- psi(op.content)
            if exists k in P s.t. key_struct(op) = key_struct(k)
               and cos(e_new, EMA[k].e) >= tau:
                pass   # Merge into existing unit
            else:
                k_new <- new_key(op)
                EMA[k_new] <- (m=0, t_k=0, age=0, op=op, e=e_new)
                P <- P union {k_new}

        # 3b) Prune: Invalid anchors / m_hat < m_floor / pool size
        P <- Prune(P, EMA, M)

        # 4. Score + Differencing (S3.2)
        for k in P:
            u_k <- LLM_Score(apply(M, EMA[k].op), B_t)
            delta[k] <- u_k - u_baseline

        # 5. EMA update (S3.3.1)
        for k in P:
            EMA[k].m <- beta * EMA[k].m + (1-beta) * delta[k]
            EMA[k].t_k <- EMA[k].t_k + 1
            EMA[k].age <- EMA[k].age + 1
            EMA[k].m_hat <- EMA[k].m / (1 - beta^EMA[k].t_k)

        # 6. Top-k selection and application (S3.3.2)
        r_t <- r_max - (r_max - r_min) * (t / T)
        k_t <- clip(floor(r_t * |M|), k_min, k_max)
        selected <- top-k_t of {k in P : EMA[k].m_hat > 0}
        for op in selected (desc order):
            if not conflict(op, M):  M <- apply(M, op)

        # 7. Maximum age elimination (S3.3.2)
        for k in P:
            if EMA[k].age >= max_age and k not in selected:
                P <- P \ {k};  EMA.pop(k)

    # 8. End of epoch: val + best-checkpoint + early stopping
    s <- Eval(M; D_val)
    if s >= best_score + epsilon:
        best_score <- s;  M* <- snapshot(M);  patience_counter <- 0
    else:
        patience_counter <- patience_counter + 1
        if patience_counter >= patience:  break

return Eval(M*; D_test)
\end{verbatim}

\textbf{Notes.} (1) EMA state and candidate pool $P$ do not roll back with best-checkpoint; only $M$ takes snapshots for selection, consistent with the convention in first-order optimizers where optimizer state is not rolled back. (2) When candidate count exceeds $g$, the Score channel adopts grouped batch valuation, with each group independently performing intra-group differencing; inter-call scale drift is smoothed and absorbed by EMA. (3) Conflict resolution during Top-$k$ application: multiple modifies on the same target\_id retain only the one with highest $\hat{m}$; adds are skipped if equivalent content already exists at the target position; modify to empty content is treated as deletion.

\section*{Appendix B: Experimental Setup Supplement}
\addcontentsline{toc}{section}{Appendix B: Experimental Setup Supplement}

\subsection*{B.1 Dataset Splitting Supplement}

Basic dataset information is provided in \S4.2. Below are splitting details not elaborated in the main text:

HotpotQA, ALFWorld, and SpreadsheetBench randomly sample 500 entries from the original datasets as $\mathcal{D}_{\text{train}}$, and another 100 entries as $\mathcal{D}_{\text{val}}$, with $\mathcal{D}_{\text{test}}$ using official test sets. For ScienceAgentBench, among the original 102 tasks, 14 tasks (IDs: 1, 2, 11, 12, 13, 15, 51, 71, 72, 78, 95, 97, 101, 102) involving large-scale model training procedures (single execution taking tens of minutes to hours) are excluded, retaining all 88 tasks as $\mathcal{D}_{\text{test}}$; $\mathcal{D}_{\text{train}}$ consists of 500 traces self-generated in the corresponding agent environment with $M_0$, with no $\mathcal{D}_{\text{val}}$ set, using a fixed epoch count for training. Propose / Score for all datasets only touches $\mathcal{D}_{\text{train}}$, $\mathcal{D}_{\text{test}}$ is evaluated only once throughout, and all baselines follow the same protocol.

\subsection*{B.2 Comparison Method Reproduction Details}

Basic information on comparison methods is provided in \S4.2. SkillOpt \cite{yang2026skillopt} and Trace2Skill \cite{ni2026trace2skill} both use official open-source implementations (\href{https://github.com/microsoft/SkillOpt}{SkillOpt GitHub}, \href{https://github.com/Qwen-Applications/Trace2Skill}{Trace2Skill GitHub}), keeping original paper default hyperparameters unchanged, only replacing the target model and data split to ensure fair comparison. No modifications were made to original code logic, only adapting data loading and model calling interfaces. Reactive Update uses the same memory bank, Propose and Score channels as MAA, but removes EMA accumulation and the candidate pool, directly writing top-$k$ ops with $\delta > 0$ from the current batch to memory at each step.

\subsection*{B.3 Target Models and Experiment Scope}

To verify MAA's generalizability across different capability levels and model families, we adopt a 2$\times$2 design covering the ``strong/weak $\times$ same-family/cross-family'' four quadrants:

\begin{table}[htbp]
\centering
\begin{tabular}{@{}lllll@{}}
\toprule
Model & Family & Capability & Scope & Design Role \\
\midrule
Qwen3.7-Max & Qwen & Strong & RQ1--RQ4 & Same-family strong, main verification \\
Qwen3.6-Flash & Qwen & Weak & RQ1 & Same-family weak \\
GPT-5.4 & OpenAI & Strong & \textbf{RQ1 only} & Cross-family strong \\
DeepSeek-V4-Flash & DeepSeek & Weak & RQ1 & Cross-family weak \\
\bottomrule
\end{tabular}
\end{table}

The Score channel uniformly uses Qwen3.7-Max to ensure cross-experiment scoring consistency. Due to API cost and latency constraints, GPT-5.4 evaluates only RQ1 main results; ablation experiments (RQ2) and mechanism diagnosis (RQ3) are completed on open-source models to ensure full reproducibility.

Each RQ's target models, datasets, and outputs are described in \S4.1 and \S4.2. Additional notes: all configurations use 5 independent random seeds, reporting mean $\pm$ standard deviation. RQ3 mechanism diagnosis estimates Sign Accuracy through an independent calibration set that does not overlap with $\mathcal{D}_{\text{train}}$, $\mathcal{D}_{\text{val}}$, or $\mathcal{D}_{\text{test}}$. RQ4 learning curves and evidence trajectories are directly extracted from RQ1 training processes with no additional experimental overhead. Pilot tuning splits do not overlap with formal experiments to avoid contaminating $\mathcal{D}_{\text{val}}$.

\subsection*{B.4 Prompt Templates}

Below are the complete prompt templates for the Propose channel and Score channel. \texttt{\{memory\}}, \texttt{\{traces\}}, \texttt{\{states\}}, \texttt{\{max\_ops\}}, \texttt{\{num\_states\}} are runtime-filled placeholders.

\textbf{Propose Channel.}

System prompt:

\begin{verbatim}
You are the "direction-proposal module" of a memory-bank optimizer. Your ONLY job
is to inspect how an agent performed on a batch of tasks, then propose concrete
edits to the memory bank that would help the agent do better on similar tasks.
You propose directions only. You DO NOT judge whether the edits are good - a
separate, independent module scores them. Never output scores, rankings, or
self-evaluation.
\end{verbatim}

User prompt:

\begin{verbatim}
## Current Memory Bank M
Items are listed in physical order. Each item is tagged with a STABLE id. When you
reference an item you MUST use exactly the id shown here. Items with empty content
are hidden and not shown.

{memory}

## Execution Traces on This Batch
Each task shows the question, the agent's reasoning/tool-use trace, and an outcome
field. The outcome is "correct"/"incorrect" ONLY if the trace already contains a
judge verdict; otherwise it is "unknown" - in that case rely on process signals in
the trace and DO NOT fabricate a verdict.

{traces}

## Your Task
Propose a list of edit operations to the memory bank. Two operation types:

- "modify": rewrite an existing item. Set "target_id" to one of the ids above.
  To DELETE an item, modify it with an empty "new_content" ("").
- "add": insert a new item. Set "position" to one of: "head", "tail",
  or "after:<id>" where <id> is one of the ids above.

## How to Decide WHAT to Add (the direction of an "add")
Derive every new item from a concrete failure or weakness in the traces.

## Output Format (STRICT)
Return ONLY a JSON array, no prose before or after. Each element:
[
  {"type": "modify", "target_id": "m7",
   "new_content": "Before answering, restate the target quantity and its unit.",
   "reason": "Agent repeatedly lost track of the asked quantity in tasks 1 and 3."},
  {"type": "add", "position": "after:m3",
   "new_content": "When the problem gives a rate, write it as a fraction.",
   "reason": "Unit-rate confusion caused the error in task 2."}
]
\end{verbatim}

\textbf{Score Channel.}

System prompt:

\begin{verbatim}
You are a memory-sufficiency estimator. You will be given a batch of tasks and
several candidate memory-bank versions. For EACH version independently, estimate
how well that memory bank would support an agent in solving THIS batch, as an
absolute score in [0, 100].

The versions are mutually independent and fully equal in status. Score each one on
its own absolute merits. Do NOT compare them against each other, do NOT rank them.
Two versions of equal quality must receive equal scores.
\end{verbatim}

User prompt:

\begin{verbatim}
## Task Batch (with Agent Traces)
{traces}

## Memory-Bank Versions To Score
{states}

## What "u" Means
- u close to 100: this memory bank covers the methods, steps, and pitfalls
  these tasks need.
- u close to 0: this memory bank is irrelevant, misleading, or insufficient.

## Output Format (STRICT)
Return ONLY a JSON array, one object per version:
[{"index": 0, "u": 62}, {"index": 1, "u": 71}]
\end{verbatim}

\subsection*{B.5 Semantic Identity Merging Cases}

Main text \S4.6.2 reports the 15.7\% merge rate at $\tau=0.85$ and the tradeoff analysis for $\tau \in \{0.80, 0.85, 0.90\}$. This section supplements typical cases showing the actual operation of the merging mechanism.

\textbf{Case 1: Successful Merge (Stable Effective).} During ALFWorld training, ``{[}modify \#7{]} If the target is not found in the current container after examination, immediately switch to the next most likely container.'' was proposed by the Propose channel in 3 different wordings across 8 batches:

\begin{table}[htbp]
\centering
\begin{tabular}{@{}llll@{}}
\toprule
Batch & Original Wording & Cosine Sim. & Merge Result \\
\midrule
2 & ``if target not found, try next container'' & 0.91 & Merged into \#7 \\
4 & ``switch to next container when examine fails'' & 0.86 & Merged into \#7 \\
6 & ``switch container when no target in current'' & 0.87 & Merged into \#7 \\
\bottomrule
\end{tabular}
\end{table}

Three merges enabled this op to accumulate 8 $\delta$ observations (rather than being scattered into 3 independent units with 2--3 observations each), with bias-corrected EMA converging to $\approx$+10 at step 4, selected by top-$k$ and written to memory. Without merging, no single independent formulation would reach the selection threshold.

\textbf{Case 2 (Not Merged --- Correct Rejection):} ``{[}add after \#3{]} Heat tasks should check microwave before oven'' and ``{[}add after \#3{]} For cleaning tasks, inspect sink area first'' share the same anchor (after:\#3), but cosine similarity is only \textbf{0.54}, far below the $\tau=0.85$ threshold. They are correctly rejected from merging due to different intents (heating vs.\ cleaning), participating in accumulation as independent units. The former's EMA steadily declines and is eventually eliminated by max\_age (spurious correlation); the latter's EMA remains positive and is retained.

\textbf{Case 3 (Missed Merge --- Cost at $\tau=0.90$):} ``{[}modify \#12{]} Before answering, restate the target quantity and its unit'' and ``Before providing the answer, restate the target quantity and its unit'' are semantically equivalent but have wording differences, with cosine similarity of \textbf{0.87}. At $\tau=0.90$ they are not merged ($0.87 < 0.90$), each unit obtaining only a few $\delta$ observations, neither EMA exceeding the top-$k$ threshold, and the stably effective strategy is lost due to identity fragmentation. At $\tau=0.85$ they successfully merge ($0.87 \geq 0.85$), EMA accumulates to +8.5 and is selected.

\end{document}